\documentclass[lettersize,journal]{IEEEtran}
\usepackage{iftex}
\ifPDFTeX
  \usepackage[T1]{fontenc}
\else
  \usepackage{fontspec}
\fi
\usepackage{cite}
\let\citep\cite
\author{Anonymous authors}

\usepackage{amsmath,amsfonts,bm}

\def\eqref#1{equation~\ref{#1}}

\def\1{\bm{1}}

\DeclareMathAlphabet{\mathsfit}{\encodingdefault}{\sfdefault}{m}{sl}
\SetMathAlphabet{\mathsfit}{bold}{\encodingdefault}{\sfdefault}{bx}{n}

\usepackage{booktabs}
\usepackage[hidelinks]{hyperref}
\usepackage{url}
\usepackage{graphicx}
\usepackage{stfloats}

\usepackage{amsfonts,amssymb}
\usepackage[ruled]{algorithm2e}
\usepackage{colortbl} 
\usepackage{xcolor}
\usepackage{bbm}
\usepackage{amsthm}
\usepackage{amssymb}
\usepackage{amsfonts}
\usepackage{amsmath}
\usepackage{subfigure}
\usepackage{mathrsfs}
\usepackage{framed}
\usepackage{epsfig}
\usepackage{rotating, multirow}
\usepackage{epstopdf}
\usepackage{float}
\usepackage{diagbox}
\usepackage{balance}
\usepackage{arydshln}
\usepackage{mathtools} 
\usepackage{breqn}

\newtheorem{proposition}{Proposition}

\newcommand\defeq{\stackrel{\mathclap{\small{\text{def}}}}{=}}
\newcommand{\cmmnt}[1]{}
\allowdisplaybreaks[4]

\title{Reorienting Adversarial Optimization for \\ Transferable Vision-Language Attacks}
\author{Peng-Fei Zhang\\
Tongji University\\
\texttt{mima.zpf@gmail.com} \\
\And
Jianfeng Qu\\
Renmin University of China\\
\texttt{jfqu@ruc.edu.cn} \\
\AND
Hengtao Shen\\
Tongji University\\
\texttt{huang@itee.uq.edu.au}
}

\author{Kaiwen Shen, Peng-Fei Zhang, Jianfeng Qu, Guanhua Ye, Yang Yang and Heng Tao Shen, \textit{IEEE Fellow}
\thanks{This work is partially supported by the National Natural Science Foundation of China (62606367) and Basic Research Funds for Central University (21002150073).}
\thanks{Kaiwen Shen and Yang Yang are with the University of Electronic Science and Technology of China, email: ckevin.shen2001@gmail.com and yang.yang@uestc.edu.cn.}
\thanks{Peng-Fei Zhang and Heng Tao Shen are with the School of Computer Science and Technology, Tongji University, email: mima.zpf@gmail.com and shenhengtao@tongji.edu.cn.}
\thanks{Jianfeng Qu is with the School of Smart Governance, Renmin University of China, email: jfqu@ruc.edu.cn.}
\thanks{Guanhua Ye is with the School of Computer Science, Beijing University of Posts and Telecommunications, email: g.ye@bupt.edu.cn.}
}

\begin{document}

\maketitle

\begin{abstract}
Adversarial attacks are crucial for evaluating the robustness of vision-language pre-trained (VLP) models. However, existing attacks often suffer from limited transferability across unseen models, restricting their effectiveness in assessing cross-model robustness. A key limitation is insufficient exploration of the adversarial search space, which may lead to overfitting to surrogate models and poor generalization to unseen models. To address this limitation, we propose Reorienting Adversarial Optimization (RAO), a unified framework that enhances attack transferability through diversified adversarial search across visual and textual modalities. For image attacks, RAO exploits historical optimization trajectories to broaden the directional search space and explore diverse optimization directions, enabling the optimization process to escape potentially suboptimal trajectories. For text attacks, RAO comprehensively explores candidate word substitutions and jointly evaluates their effectiveness in disrupting global and local cross-modal semantic alignment. By reorienting adversarial optimization beyond locally preferred updates, RAO mitigates reliance on surrogate-specific optimization patterns and promotes more transferable perturbations across modalities. Extensive experiments across diverse benchmarks, VLP models, and downstream tasks, supported by theoretical analysis, demonstrate that RAO achieves superior attack transferability compared with state-of-the-art methods. 
The code is released at \url{https://github.com/RAO-Attack/RAO}.
\end{abstract}

\section{Introduction}
Vision-language pre-trained (VLP) models learn cross-modal relationships and transferable representations from large-scale image-text data, serving as a foundation for various downstream tasks, including image-text retrieval, image captioning, and visual question answering \cite{radford2021learning,li2022blip,li2021align,yang2022vision}. However, these models remain vulnerable to adversarial perturbations, where subtle modifications to input images or texts can disrupt cross-modal alignment and mislead model predictions \cite{zhang2022towards,lu2023set,yin2023vlattack,zhang2026data,zhang2024effective}. Investigating such vulnerabilities is therefore essential for assessing the robustness of VLP models and improving their reliability in practical applications.

In practical scenarios, crafting model-specific adversarial examples for each target model incurs substantial computational costs, while the architectures and parameters of target models are often inaccessible. Transfer-based attacks address this challenge by generating adversarial examples on accessible surrogate models and transferring them to unseen target models, providing a practical approach to black-box robustness evaluation \cite{goodfellow2014explaining,fang2022learning,huang2024enhancing,xu2023backpropagation,zhang2024universal}. However, discrepancies in model architectures, training objectives, and cross-modal representations often hinder the transferability of adversarial perturbations. 

To overcome this limitation, existing studies have explored various approaches. For image-based attacks, common strategies include cross-modal interaction disruption, data augmentation, and layer-wise feature divergence. For example, SGA \cite{lu2023set} and SimVLA \cite{ren2026success} leverage data augmentation to enhance set-level cross-modal guidance, while DRA \cite{gao2025boosting} and SA-AET \cite{jia2025semantic} diversify adversarial examples through trajectory-guided augmentation. Other approaches \cite{li2026towards,liu2026hqa} incorporate semantic guidance to steer adversarial perturbations toward semantically distinct representations. Despite these advances, existing image-based attacks often rely on current gradient directions or explore perturbations within constrained regions along optimization trajectories, potentially overlooking more effective update directions and leading to suboptimal optimization paths. For text-based attacks, existing methods commonly prioritize word substitutions based on word importance \cite{lu2023set,yin2023vlattack,jia2025semantic}. However, word importance does not necessarily reflect substitution effectiveness, as less salient words may yield stronger adversarial perturbations when replaced. Furthermore, evaluating substitutions primarily through global image-text similarity may overlook fine-grained semantic correspondences between visual regions and textual concepts. Consequently, both image- and text-based attacks may suffer from insufficient exploration of the adversarial search space, limiting the discovery of effective perturbations and their transferability to unseen models.

In this paper, we propose Reorienting Adversarial Optimization (RAO), a simple yet effective attack framework that enhances adversarial transferability through more comprehensive exploration of the adversarial search space. For image attacks, RAO leverages historical optimization trajectories to construct and evaluate diverse candidate directions, enabling optimization to deviate from potentially suboptimal paths and mitigate the risk of convergence to local optima.  Beyond extensive empirical evaluations, we provide theoretical analysis to further elucidate the effectiveness of this optimization mechanism. For the text modality, rather than restricting substitutions to the most salient word, RAO moves beyond saliency-driven word substitution by systematically exploring alternative replacements and jointly evaluating their effectiveness in disrupting global and local cross-modal semantic alignment. Extensive experiments demonstrate that RAO consistently outperforms state-of-the-art methods, substantially improving adversarial transferability across both image and text modalities.

The main contributions of this paper are as follows:
\begin{enumerate}
\item Reorienting Adversarial Optimization (RAO) is proposed to address insufficient exploration of the adversarial search space to enhance cross-model attack transferability across visual and textual modalities.
\item For image attacks, RAO exploits historical optimization trajectories to construct and evaluate diverse candidate directions, mitigating the risk of suboptimal optimization paths. For text attacks, RAO systematically explores alternative word substitutions and jointly evaluates their effectiveness against global and local cross-modal semantic alignment.
\item Extensive experiments demonstrate that RAO consistently outperforms state-of-the-art methods, achieving the highest average black-box attack success rates across diverse tasks, validating its superior adversarial transferability.
\end{enumerate}

{\color{black}
\section{Related Work}\label{Rel}

\subsection{Vision-Language Pre-trained Models}
Vision-language pre-training (VLP) learns transferable multimodal representations from large-scale image-text data, providing a foundation for tasks such as image-text retrieval, image captioning, and visual question answering. CLIP~\citep{radford2021learning} uses separate image and text encoders to learn aligned representations; ALBEF~\citep{li2021align} performs cross-modal alignment before multimodal fusion; BLIP~\citep{li2022blip} improves pre-training data through caption generation and filtering; and TCL~\citep{yang2022vision} combines cross-modal, intra-modal, and local-global contrastive learning to enhance representations. Differences in their architectures and training objectives make these models representative choices for studying cross-model adversarial transferability.

\subsection{Adversarial Attacks}

Adversarial attacks have been extensively investigated to assess the robustness of deep learning models by introducing subtle perturbations that mislead model predictions~\citep{madry2018towards,zhang2026hierarchical,dong2018boosting,kurakin2016adversarial,zhang2023proactive,fang2025one}. Images and texts exhibit distinct data characteristics, with images represented as continuous-valued signals  and texts composed of discrete tokens. Accordingly, image-based attacks typically employ gradient-based optimisation to generate continuous perturbations \cite{szegedy2014intriguing,moosavi2017universal,wang2021boosting,xu2024unified}, whereas text-based attacks primarily rely on word substitutions to construct adversarial examples \cite{liang2018deep,jin2020bert,liu2023aliasing,ren2019generating}. In the context of vision-language models, transfer-based attacks have attracted considerable attention, as they enable adversarial examples crafted on surrogate models to be transferred to unseen target models. To enhance transferability, existing studies have developed various approaches tailored to the characteristics of visual and textual modalities.

For image-based attacks, existing studies have explored various approaches to enhance adversarial transferability. One prominent direction leverages data augmentation to exploit cross-modal interactions and improve the generalizability of adversarial perturbations. For instance, SGA~\citep{lu2023set} employs set-level guidance across augmented image-text pairs to strengthen cross-modal adversarial optimization, while SimVLA~\citep{ren2026success} simplifies the attack pipeline to facilitate transferable attacks. Another line of research investigates the contributions of different network layers to adversarial transferability. Specifically, existing methods~\citep{yin2023vlattack,liu2026hqa,zhang2025maa} exploit low-level feature representations to disrupt fine-grained cross-modal alignment. In addition, semantic-guided approaches~\citep{liu2026hqa,li2026towards} enhance attack transferability by increasing the discrepancy between adversarial and original representations while steering perturbations toward mismatched semantics. More recently, studies have explored adversarial perturbation diversification by leveraging information from optimization trajectories. In particular, DRA~\citep{gao2025boosting} and SA-AET~\citep{jia2025semantic} incorporate clean images and intermediate adversarial examples to diversify perturbations. For text-based attacks, existing approaches primarily rely on word-level substitutions to generate adversarial texts while preserving linguistic plausibility. A common strategy estimates word importance by measuring the influence of individual words on model predictions and subsequently replaces selected words with semantically similar alternatives~\citep{lu2023set,yin2023vlattack,zhang2025modality}. To improve attack effectiveness, subsequent studies further incorporate cross-modal semantic relationships into textual adversarial optimization~\citep{liu2026hqa,li2026towards}. These methods leverage semantic information to guide candidate generation and substitution evaluation, seeking to disrupt image-text alignment while maintaining textual consistency.

\par}

\section{Methodology}

\subsection{Preliminaries}

We consider a transfer-based black-box attack setting, where the target model and its downstream tasks are unknown. To generate adversarial examples transferable to unseen targets, we employ an accessible surrogate VLP model $F_S$, consisting of an image encoder $f_I$ and a text encoder $f_T$. Given a set of $B$ image-text pairs $\{(x_i,t_i)\}_{i=1}^{B}$, the objective is to generate adversarial images $x_i'=x_i+\delta_i$ and/or adversarial texts $t_i'$ to disrupt cross-modal alignment on unseen target models. The image perturbations $\Delta=\{\delta_i\}_{i=1}^{B}$ are constrained by $\|\delta_i\|_\infty\leq\epsilon_I$, while text perturbations are limited to at most $\epsilon_T$ word substitutions.

\begin{figure*}
\color{black}%
\color{black}
\center
  \includegraphics[width=0.9\textwidth]{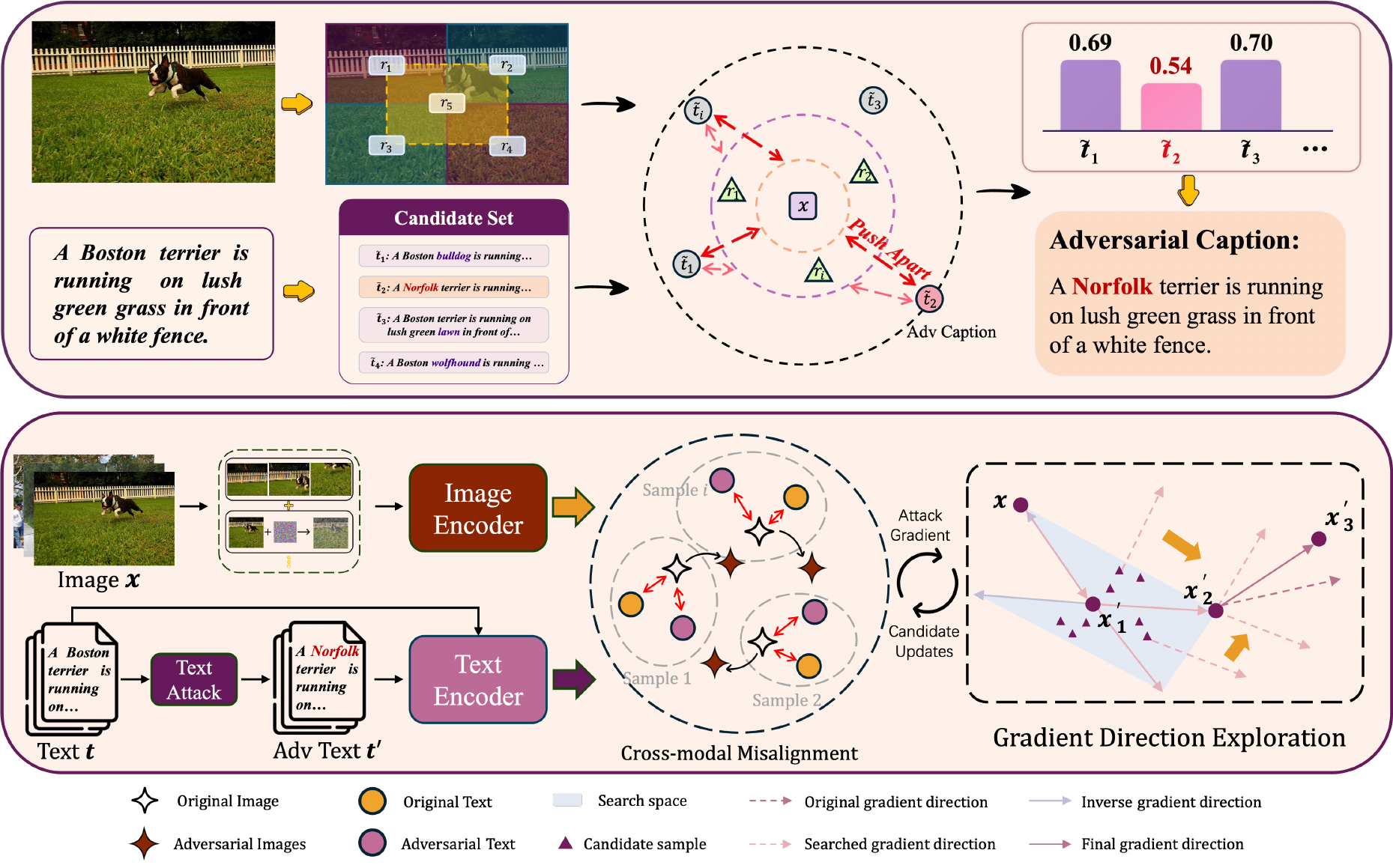}
   \caption{Overview of Reorienting Adversarial Optimization (RAO). Top: text attacks jointly evaluate candidate substitutions using global image--text similarity and average region--text similarity to disrupt cross-modal alignment at both levels. Bottom: image attacks combine data augmentation, Cross-modal Misalignment, and Gradient Direction Exploration. Augmented views provide diverse visual inputs, while Cross-modal Misalignment decreases similarities to matched clean and adversarial texts and increases similarities to mismatched adversarial texts. Gradient Direction Exploration combines historical gradients to construct candidate directions and uses the cross-modal objective to select effective candidates for perturbation updates.}
  \label{fig:rao}
\end{figure*}

\textcolor{black}{Figure~\ref{fig:rao} presents the overall framework of RAO. Adversarial texts are first generated through global-local candidate evaluation and then used together with the original texts to guide image perturbation optimization.}

\subsection{Image Attack}
\noindent\textbf{Cross-modal Misalignment.} 
To disrupt cross-modal semantic alignment, we optimize adversarial image perturbations by reducing the similarity between matched image-text representations while encouraging alignment with mismatched representations. To improve the generalizability of adversarial perturbations, we further incorporate diverse input transformations into cross-modal adversarial optimization. Specifically, we construct an augmentation set $\mathcal A$ comprising random cropping and resizing, Gaussian noise injection, random translation, and random resizing with padding. These transformations introduce complementary variations in spatial configurations and visual appearance, encouraging adversarial perturbations to remain effective across diverse visual conditions.

Let $\mathcal P_i$ and $\mathcal N_i$ denote the sets of matched and mismatched text indices for image $x_i$, respectively. Given a transformation $a$ sampled from $\mathcal A$, the attack objective is formulated as
\begin{equation}\label{eq:image-contrastive}
\begin{aligned}
\mathcal J_I(\Delta)
=\frac{1}{B}\sum_{i=1}^{B}
\mathbb E_{a\sim\mathcal A}\Bigg[
&-\sum_{j\in\mathcal P_i}
s\big(f_I(a(x_i')),f_T(t_j)\big)\\
&-\sum_{j\in\mathcal P_i}
s\big(f_I(a(x_i')),f_T(t_j')\big)\\
&+\sum_{j\in\mathcal N_i}
s\big(f_I(a(x_i')),f_T(t_j')\big)
\Bigg],
\end{aligned}
\end{equation}
where $s(\cdot,\cdot)$ denotes cosine similarity. The objective is maximized with respect to $\Delta$ under the perturbation constraints defined in the preliminaries.

By reducing the alignment between adversarial images and their matched original and adversarial texts while encouraging similarity with mismatched text representations, the objective aims to disrupt cross-modal semantic correspondence. Incorporating diverse transformations further encourages perturbations to maintain their adversarial effectiveness across different input variations. However, conventional gradient-based optimization primarily relies on the current gradient direction, which may lead to suboptimal optimization trajectories and overfitting to the surrogate model. To address these limitations, RAO exploits historical optimization information to explore a broader range of potential optimization directions, facilitating the discovery of more effective adversarial perturbations. The image attack pipeline is illustrated in the lower part of Fig.~\ref{fig:rao}.

\noindent\textbf{Gradient Direction Exploration.}
To diversify gradient-based optimization, we retain the input gradients evaluated at the current perturbation in each optimization step. These current and historical gradients are then combined using randomly sampled coefficients to construct diverse candidate directions.

Specifically, at optimization step $k$, we collect historical gradients together with the current gradient, forming a gradient set $\{g(x+\delta_0),g(x+\delta_1),\ldots,g(x+\delta_{k-1})\}$, where $g(\cdot)$ denotes the gradient of the attack objective with respect to the input image. For each sampling scheme, we use these gradients to construct a set of $D$ candidate directions, denoted as $\mathbb G_k=\{\hat g_k^1,\hat g_k^2,\ldots,\hat g_k^D\}$. Each candidate direction $\hat g_k^d$ is generated through a weighted combination of the collected gradients:
\begin{equation}\label{eq:image-direction-combination}
\hat g_k^d =
\sum_{q=1}^{k}\mu_q^{k,d}g(x+\delta_{q-1}),
\quad
\text{s.t.}\quad
\sum_{q=1}^{k}\mu_q^{k,d}=1,
\end{equation}
where $\mu_q^{k,d}$ denotes the combination coefficient assigned to the gradient from step $q$ when constructing the $d$-th candidate direction. For simplicity, the image index is omitted in the following direction construction and perturbation update equations.

To enhance directional diversity, we sample combination coefficients from both a uniform distribution $\mathcal U(0,1)$ and a standard normal distribution $\mathcal N(0,1)$, followed by normalization. Uniform sampling produces nonnegative combinations of historical gradients, whereas Gaussian sampling permits both positive and negative contributions, allowing the constructed directions to extend beyond the convex hull of historical gradients. By combining these two sampling strategies, RAO explores a more diverse range of optimization directions.

\noindent\textbf{Candidate Direction Selection.}
To identify promising optimization directions, we evaluate the constructed candidates based on their cross-modal adversarial effectiveness. Specifically, candidate directions are retained if their corresponding perturbations improve the attack objective over the unperturbed baseline:
\begin{equation}\label{eq:image-direction-selection}
\begin{aligned}
&\mathbb G_k^{m+}
=
\Big\{
\hat g_k^{m,d}\in\mathbb G_k^m
\;\Big|\;
\mathcal J_I(\hat\Delta_k^{m,d})
>
\mathcal J_I(\mathbf 0)
\Big\},\\
&\text{s.t.}\quad
\hat\delta_k^{m,d}
=
\delta_{k-1}+\lambda\hat g_k^{m,d},
\quad m\in\{1,2\},
\end{aligned}
\end{equation}
where $\lambda$ denotes the exploration step size, $m=1$ and $m=2$ index the search spaces constructed using uniform and Gaussian coefficient sampling, respectively, and $\hat\Delta_k^{m,d}$ collects the candidate perturbations across the batch. Each space is constructed using Eq.~\ref{eq:image-direction-combination}. The baseline $\mathcal J_I(\mathbf 0)$ is evaluated without image perturbations.

\noindent\textbf{Perturbation Learning.}
The perturbations corresponding to the selected candidate directions from both search spaces, together with the current perturbation, form the perturbation set $\tilde{\Delta}_k$. At each perturbation location, we perform gradient-based optimization to obtain a corresponding update, and aggregate these updates to refine the current perturbation:
\begin{equation}\label{eq:image-perturbation-update}
\begin{aligned}
&\delta_k
=
\operatorname{Clip}_{\epsilon_I}
(
\delta_{k-1}
+
\sum_{\tilde\delta\in\tilde\Delta_k}
\epsilon_k^{(\tilde\delta)}
),\\
&\text{s.t.}\quad
\epsilon_k^{(\tilde\delta)}
=
\alpha\operatorname{sign}
\left(
\nabla_{\tilde\delta}
\mathcal J_I
\right),
\end{aligned}
\end{equation}
where $\epsilon_k^{(\tilde\delta)}$ denotes the optimization update obtained at perturbation location $\tilde\delta$, and $\alpha$ is the step size. The current-point input gradients are retained as historical optimization information for subsequent direction exploration.

\subsubsection{Theoretical Analysis}\label{Theo_anan}
Adversarial transferability is influenced by multiple factors. Our method seeks to explore a broader space of update directions by leveraging both current and historical gradient information, thereby reducing the risk of overfitting and becoming trapped in local optima. A selection mechanism is incorporated to ensure that the generated updates remain effective. Beyond this empirical motivation from optimization dynamics, we provide theoretical analysis showing that our approach reduces interactions between perturbation units, which have been associated with adversarial transferability.

Wang \textit{et al.} \cite{wang2020unified} demonstrated that interactions between perturbation units are negatively correlated with adversarial transferability. A common approach to quantify such interactions is via a Shapley value–based interaction index. Specifically, let $\mathcal{D}$ denote the set of all units (or dimensions) within a perturbation, and let $S \subseteq \mathcal{D}$ be a subset. Define $S_{i j}=\{i, j\}$ as a combined unit of $i \ \text{and} \ j$. The set $\mathcal{D}^{\prime} = (\mathcal{D} \setminus \{i,j\}) \cup \{S_{i j}\}$ represents the perturbation units in which $i \ \text{and} \ j$ are treated jointly as a combined unit $S_{i j}$. Let $\phi (\cdot)$ denote the Shapley value. The interaction between perturbation units $i \ \text{and} \ j$ is then defined as:
\begin{equation}
    I_{i j}(\delta) = \phi(S_{i j}|\mathcal{D}^{\prime}) - \left[\phi(i|\mathcal{D}\setminus\{j\}) + \phi(j|\mathcal{D}\setminus\{i\})\right], \label{eq:interaction}
\end{equation}
where $\phi(S_{i j}|\mathcal{D}^{\prime})$ represents the contribution of $S_{i j}$. $\phi(i|{\mathcal{D}\setminus\{j\}})$ and $\phi(j|\mathcal{D}\setminus\{i\})$ denote the contributions of units $i$ and $j$ when the other is absent, respectively. The interaction strength is indicated as $|I_{i j}(\delta)|$. Positive and negative interaction values indicate synergistic and antagonistic effects between perturbation units, respectively.

Based on this interaction measure, we present the following proposition. For theoretical tractability, we consider gradient updates before the sign operation and omit clipping.
\begin{proposition}\label{pro:2}
Consider the perturbations generated by the gradient aggregation mechanism of RAO, expressed as
$\delta_k^{\mathrm{RAO}}
=\alpha\sum_{r=1}^{k}g_r$,
where
\begin{equation}\label{eq2-1}
g_k=
\begin{cases}
\displaystyle
\sum_{\tilde{\delta}\in\tilde{\Delta}_k}
\nabla_{\tilde{\delta}}
\mathcal J_I(\tilde{\delta}),
& k>1,\\[8pt]
g(x), & k=1.
\end{cases}
\end{equation}

The adversarial perturbation generated by a general multi-step attack is 
$\delta_k^{\mathrm{msa}}
=\alpha\sum_{r=0}^{k-1}
\nabla_{\delta}\mathcal J_I(\delta_r^{\mathrm{msa}})$. The interaction of perturbations generated by the proposed method is smaller than that of the general multi-step attack,
\emph{i.e.} $\mathbb E_{a,b}
[I_{ab}(\delta_k^{\mathrm{RAO}})]
\leq
\mathbb E_{a,b}
[I_{ab}(\delta_k^{\mathrm{msa}})]$. (Detailed proof can be found in Appendix~\ref{sec:proof_mi}.)

\end{proposition}

\subsection{Text Attack}

Existing text-based attacks commonly prioritize word substitutions based on word importance and evaluate their effectiveness through global image-text similarity. However, word importance does not necessarily reflect substitution effectiveness, while global representations may overlook fine-grained semantic correspondences between textual concepts and local visual regions. To address these limitations, RAO explores candidate word substitutions more comprehensively and jointly considers global and local visual semantics to guide their selection.

Specifically, given a text sequence containing $L$ words, we generate candidate substitutions at different word positions using Counter-Fitting (CF) neighbors with cosine similarity greater than $\tau$. A pre-trained BERT model~\citep{li2020bert} is used only when the original word is absent from the CF vocabulary. The resulting candidate set is defined as
$\tilde{T}=\bigcup_{j=1}^{L}\{\tilde{t}_j^1,\tilde{t}_j^2,\ldots,\tilde{t}_j^{l_j}\}$,
where $l_j$ denotes the number of valid substitutions for the $j$-th word. Here, $\tilde{t}_j^l$ denotes the complete candidate text obtained by replacing the $j$-th word in the original text with its $l$-th candidate substitution. Given an image $x$, we extract $K$ local regions $\mathcal R(x)=\{r_1,\ldots,r_K\}$ to capture fine-grained visual semantics. Specifically, we adopt a five-region configuration ($K=5$) comprising four corner crops and one central crop, each covering 50\% of the image width and height. This configuration provides spatial coverage with overlapping contextual information, enabling the evaluation of local cross-modal correspondences across different image regions. We then evaluate candidate substitutions by jointly considering their discrepancies with the global image representation and the representations of local visual regions. The adversarial text is selected by maximizing the global-local semantic discrepancy:
\begin{equation}\label{eq3-1}
\begin{aligned}
t' &= \arg\max_{\tilde{t}\in\tilde{T}}
\mathcal{L}_{GL}(x,\tilde{t}),\\
\mathcal{L}_{GL}(x,\tilde{t})
&= -S(x,\tilde{t})
-\frac{1}{K}\sum_{k=1}^{K}S(r_k,\tilde{t}),
\end{aligned}
\end{equation}
where $S(x,t)=s(f_I(x),f_T(t))$ denotes the cosine similarity between visual and textual representations. By evaluating candidate substitutions against both global and regional visual representations, RAO encourages adversarial texts to disrupt cross-modal alignment at different levels of semantic granularity, improving the exploration of effective textual perturbations.

\textcolor{black}{See Algorithm~\ref{alg1} in the Appendix for the complete procedure.}

\color{black}%
\section{Experiments}\label{Exp}
\subsection{Settings}
\noindent\textbf{Datasets, Tasks, and Models.} We comprehensively evaluate RAO and baselines on diverse datasets, including Flickr30K~\citep{plummer2015flickr30k}, MSCOCO~\citep{lin2014microsoft}, and RefCOCO+~\citep{yu2016modeling}. The evaluation covers multiple tasks, including image–text retrieval, image captioning, and visual grounding. Four widely used VLP models, namely CLIP~\citep{radford2021learning}, ALBEF~\citep{li2021align}, TCL~\citep{yang2022vision}, and BLIP~\citep{li2022blip}, are employed as source and target models. For CLIP, both ViT-B/16~\citep{dosovitskiy2020image} and ResNet101~\citep{he2016deep} backbones are considered. 

\noindent\textbf{Implementation Details.} The perturbation magnitudes are set uniformly to $\epsilon_I = 2/255$ for images and $\epsilon_T = 1$ for text. For image attacks, we employ Projected Gradient Descent (PGD)~\citep{madry2018towards} with $T=10$ iterations, exploration and update step sizes $\lambda=\alpha=0.5/255$, and a batch size of 16. We generate $D=5$ candidates per search space, yielding 10 candidates across the two search spaces.

We construct augmented views using random cropping, Gaussian noise, random translation, downscaling with padding, and downscaling followed by resizing to the original resolution. The crop area ratio ranges from 0.5 to 0.8, the maximum noise standard deviation is $\sigma=0.01$, and translations are within $\pm15\%$ along each axis. Downscaling with padding uses side-length ratios of 0.75--1.0, while downscaling followed by resizing uses ratios of 0.5 and 0.75. Reflection padding is used for translation and downscaling with padding.

For text candidate generation, we use Counter-Fitting neighbors with cosine similarity greater than $\tau=0.4$. BERT is used only when the original word is absent from the Counter-Fitting vocabulary.

\noindent\textbf{Baselines.} We compare the proposed method with state-of-the-art approaches, including Co-Attack~\citep{zhang2022towards}, SGA~\citep{lu2023set}, VLATTACK~\citep{yin2023vlattack}, DRA~\citep{gao2025boosting}, SA-AET~\citep{jia2025semantic}, SimVLA~\citep{ren2026success}, HQA-VLAttack~\citep{liu2026hqa}, and SADCA~\citep{li2026towards}. 

{\color{black}
\noindent\textbf{Evaluation Metric.} The attack success rate (ASR) is employed to evaluate all methods on image-text retrieval, quantifying the proportion of adversarial examples that successfully induce incorrect predictions. For other tasks, the effect of attacks is assessed by measuring the performance drop of the target models before and after the attack.
\par}

\color{black}
\subsection{Experimental Results}

{\color{black}
\subsubsection{Cross-Model Transferability}
We evaluate the adversarial transferability of attacks across different models in image-text retrieval, which identifies relevant items for a given query by ranking their distances in the embedding space. We consider CLIP$_{\text{ViT-B/16}}$, CLIP$_{\text{ResNet101}}$, ALBEF, and TCL as source and target models, with results summarized in Table~\ref{tab_cross}. TR denotes image-to-text retrieval, while IR denotes text-to-image retrieval. Due to space constraints, only attack success rates at R@1 are reported. Black-box Avg. is the mean attack success rate over the six TR and IR results on the three target models other than the source model.

First, RAO achieves the highest average black-box attack success rate under all four source-model settings and the best performance on all 24 black-box TR and IR metrics. With CLIP$_{\text{ViT-B/16}}$, CLIP$_{\text{ResNet101}}$, ALBEF, and TCL as source models, RAO achieves black-box averages of 56.16\%, 47.48\%, 71.05\%, and 68.53\%, respectively, exceeding the strongest baseline in each setting by 8.32, 4.77, 6.04, and 7.36 percentage points. These results show that RAO generates adversarial examples with strong black-box transferability across different source models.

Second, attack success rates vary substantially across source-target model pairs, which may reflect differences in model architectures, training objectives, and learned cross-modal representations. For example, adversarial examples generated by RAO on CLIP$_{\text{ViT-B/16}}$ achieve TR and IR attack success rates of 83.52\% and 84.67\% on CLIP$_{\text{ResNet101}}$, compared with 32.01\% and 49.91\% on ALBEF. This illustrates that the effectiveness of transfer depends on the source-target model pair, even for the same attack method.

Third, transferability is asymmetric. For RAO, transferring from ALBEF to CLIP$_{\text{ViT-B/16}}$ yields TR and IR attack success rates of 62.33\% and 69.88\%, whereas the reverse direction yields 32.01\% and 49.91\%. Thus, effective transfer in one direction does not imply comparable performance in the reverse direction, highlighting the importance of evaluating attacks across multiple source-target configurations.
\par}

\begin{table*}[!t]
\centering
\caption{Attack success rate (ASR, $\%$) at R@1 for image-text retrieval on Flickr30K. {*} indicates white-box attack results; \textbf{bold} and \underline{underlining} indicate the best and second-best performance, respectively. Black-box Avg. is the mean ASR at R@1 over both retrieval directions and the three target models other than the source model.}
\label{tab_cross}
\resizebox{0.9\linewidth}{!}{
\begin{tabular}{c|l|cc|cc|cc|cc|c}
\toprule
& Target Model & \multicolumn{2}{c|}{CLIP$_{\text{ViT-B/16}}$} & \multicolumn{2}{c|}{CLIP$_{\text{ResNet101}}$} & \multicolumn{2}{c|}{ALBEF} & \multicolumn{2}{c|}{TCL} & Black-box \\\midrule
Source Model & Method & TR R@1 & IR R@1 & TR R@1 & IR R@1 & TR R@1 & IR R@1 & TR R@1 & IR R@1 & Avg. \\
\midrule
\multirow{9}{*}{CLIP$_{\text{ViT-B/16}}$} & Co-Attack & 94.64* & 96.09* & 28.59 & 40.34 & 9.91 & 24.16 & 11.17 & 25.55 & 23.29 \\
 & SGA & 99.14* & 99.19* & 41.00 & 48.13 & 12.41 & 27.69 & 16.42 & 30.74 & 29.40 \\
 & VLATTACK & 98.16* & 98.87* & 36.14 & 43.36 & 10.01 & 26.03 & 13.07 & 27.83 & 26.07 \\
 & DRA & 98.53* & 98.71* & 44.32 & 49.91 & 12.72 & 29.42 & 13.81 & 30.69 & 30.15 \\
 & SA-AET & 98.90* & 99.03* & 44.70 & 50.66 & 14.29 & 30.52 & 15.49 & 30.69 & 31.06 \\
 & SimVLA & 95.21* & 96.91* & 55.56 & 64.12 & 22.00 & \underline{42.40} & 23.18 & 43.48 & 41.79 \\
 & HQA-VLAttack & \textbf{100.00*} & \textbf{100.00*} & \underline{74.07} & \underline{77.19} & 25.13 & 41.98 & 24.66 & 44.00 & \underline{47.84} \\
 & SADCA & \underline{99.75*} & 99.90* & 65.01 & 70.84 & \underline{26.90} & 42.14 & \underline{30.77} & \underline{44.64} & 46.72 \\
 & \textbf{RAO(ours)} & \textbf{100.00*} & \underline{99.97*} & \textbf{83.52} & \textbf{84.67} & \textbf{32.01} & \textbf{49.91} & \textbf{33.83} & \textbf{53.00} & \textbf{56.16} \\
\midrule
\multirow{9}{*}{CLIP$_{\text{ResNet101}}$} & Co-Attack & 28.59 & 40.34 & 98.08* & 98.52* & 9.38 & 23.90 & 11.70 & 25.12 & 23.17 \\
 & SGA & 32.52 & 44.30 & 97.80* & 97.84* & 11.57 & 25.28 & 13.49 & 27.83 & 25.83 \\
 & VLATTACK & 32.88 & 43.23 & 99.36* & 99.28* & 3.86 & 9.33 & 8.54 & 13.71 & 18.59 \\
 & DRA & 33.87 & 46.01 & 98.72* & 98.77* & 12.41 & 26.59 & 13.80 & 29.07 & 26.96 \\
 & SA-AET & 35.09 & 46.49 & 99.62* & \underline{99.55*} & 12.13 & 27.90 & 13.49 & 28.81 & 27.32 \\
 & SimVLA & 50.60 & 61.68 & 97.96* & 99.07* & \underline{22.10} & \underline{41.40} & \underline{25.92} & \underline{44.14} & 40.97 \\
 & HQA-VLAttack & \underline{62.82} & \underline{69.46} & \underline{99.87*} & \textbf{100.00*} & 20.75 & 38.66 & 22.13 & 42.45 & \underline{42.71} \\
 & SADCA & 40.00 & 51.93 & \textbf{100.00*} & \textbf{100.00*} & 16.37 & 31.43 & 19.07 & 35.29 & 32.35 \\
 & \textbf{RAO(ours)} & \textbf{68.96} & \textbf{75.19} & \textbf{100.00*} & \textbf{100.00*} & \textbf{25.23} & \textbf{42.24} & \textbf{26.34} & \textbf{46.93} & \textbf{47.48} \\
\midrule
\multirow{9}{*}{ALBEF} & Co-Attack & 23.98 & 35.50 & 25.72 & 38.65 & 80.24* & 87.68* & 15.13 & 29.02 & 28.00 \\
 & SGA & 33.43 & 44.49 & 35.15 & 48.32 & 97.72* & 97.52* & 46.01 & 55.19 & 43.77 \\
 & VLATTACK & 36.56 & 45.78 & 35.63 & 47.17 & 96.14* & 96.09* & 41.62 & 52.81 & 43.26 \\
 & DRA & 38.93 & 48.90 & 42.53 & 50.37 & 96.98* & 96.77* & 50.79 & 61.33 & 48.81 \\
 & SA-AET & 40.12 & 50.48 & 44.19 & 52.10 & 97.39* & 98.04* & 57.06 & 64.07 & 51.34 \\
 & SimVLA & \underline{52.52} & 60.24 & 56.69 & \underline{65.88} & 94.79* & 95.09* & 63.65 & 72.88 & 61.98 \\
 & HQA-VLAttack & 52.15 & \underline{62.05} & \underline{59.64} & 65.59 & \textbf{99.79*} & \textbf{99.98*} & \underline{73.02} & \underline{77.60} & \underline{65.01} \\
 & SADCA & 47.73 & 54.22 & 50.32 & 60.65 & 99.27* & 99.28* & 56.80 & 63.83 & 55.59 \\
 & \textbf{RAO(ours)} & \textbf{62.33} & \textbf{69.88} & \textbf{67.18} & \textbf{72.45} & \underline{99.69*} & \underline{99.77*} & \textbf{74.71} & \textbf{79.76} & \textbf{71.05} \\
\midrule
\multirow{9}{*}{TCL} & Co-Attack & 28.15 & 41.66 & 30.23 & 44.46 & 22.89 & 20.90 & 79.24* & 85.62* & 31.38 \\
 & SGA & 34.39 & 44.62 & 37.53 & 48.94 & 48.82 & 60.18 & 98.33* & 99.00* & 45.75 \\
 & VLATTACK & 36.44 & 21.20 & 36.91 & 47.86 & 34.20 & 46.80 & 96.31* & 96.64* & 37.24 \\
 & DRA & 37.50 & 50.58 & 43.81 & 51.17 & 52.14 & 63.47 & 98.31* & 98.88* & 49.78 \\
 & SA-AET & 38.53 & 51.64 & 45.87 & 52.72 & 56.35 & 68.18 & 98.31* & 98.76* & 52.22 \\
 & SimVLA & 47.61 & \underline{60.15} & \underline{55.83} & \underline{65.43} & \underline{63.86} & \underline{74.17} & 97.50* & 97.59* & \underline{61.18} \\
 & HQA-VLAttack & \underline{52.39} & 59.41 & 55.43 & 62.44 & 62.88 & 71.70 & \underline{99.79*} & \textbf{99.93*} & 60.71 \\
 & SADCA & 43.80 & 52.74 & 50.83 & 60.99 & 58.71 & 67.40 & 99.37* & 99.57* & 55.75 \\
 & \textbf{RAO(ours)} & \textbf{61.84} & \textbf{67.11} & \textbf{66.41} & \textbf{70.33} & \textbf{69.13} & \textbf{76.36} & \textbf{100.00*} & \underline{99.90*} & \textbf{68.53} \\
\bottomrule
\end{tabular}
}
\end{table*}
\subsubsection{Cross-Task Transferability}

To further demonstrate the superiority of the proposed method across different tasks, we leverage adversarial examples learned from image-text retrieval to attack models for visual grounding and image captioning. Compared to image-text retrieval, these tasks consider more fine-grained relationships between images and texts: visual grounding aims to match text to corresponding regions or objects in an image, whereas image captioning generates detailed descriptions of image content, including objects, background, and their relationships. \textcolor{black}{Tables \ref{tab_trans_2} and \ref{tab_trans_3} report the cross-task transfer results. For visual grounding, RAO achieves the lowest target-model performance on all three splits, with scores of 42.42, 49.29, and 34.59 on Val, TestA, and TestB, respectively, and an average of 42.10. For image captioning, RAO achieves the lowest scores on B@4, METEOR, and ROUGE-L, indicating reduced agreement between the target model's generated captions and the reference captions. SADCA achieves slightly lower CIDEr and SPICE scores than RAO by 0.28 and 0.20 points, respectively. RAO also yields the lowest arithmetic mean of the five reported metric scores. These results support the cross-task transferability of the proposed method.}

\subsection{Ablation Study}

\begin{table*}[!t]
\color{black}%
\caption{Performance on visual grounding under different attacks on RefCOCO+. ALBEF for image-text retrieval serves as the source model, and ALBEF for visual grounding serves as the target model. ``Baseline'' refers to the target model's performance on clean data. Smaller values indicate better adversarial transferability, and \textbf{bold} highlights the best results. Avg. is the unweighted arithmetic mean over Val, TestA, and TestB.}
\label{tab_trans_2}
\centering
\resizebox{0.9\linewidth}{!}{
\begin{tabular}{lcccccccccc}
\toprule
 {} & {Baseline} & {Co-Attack} & {SGA} & {VLATTACK} & {DRA} & {SA-AET} & {SimVLA} & {HQA-VLAttack} & {SADCA} & {RAO} \\ \hline
{Val} & 58.46 & 54.26 & 50.56 & 56.67 & 49.32 & 47.44 & 45.87 & 46.43 & 45.06 & \textbf{42.42} \\
{TestA} & 65.89 & 61.80 & 57.42 & 65.37 & 56.48 & 53.27 & 50.19 & 51.90 & 53.31 & \textbf{49.29} \\
{TestB} & 46.25 & 43.81 & 36.61 & 45.28 & 40.15 & 38.58 & 36.82 & 37.29 & 35.97 & \textbf{34.59} \\
\midrule
{Avg.} & 56.87 & 53.29 & 48.20 & 55.77 & 48.65 & 46.43 & 44.29 & 45.21 & 44.78 & \textbf{42.10} \\
\bottomrule
\end{tabular}
}
\end{table*}

\begin{table*}[!t]
\color{black}%
\caption{ Performance in the image captioning task under various attacks on MSCOCO. ALBEF for image-text retrieval serves as the source model, while BLIP built for image captioning is used as the target model. ``Baseline'' refers to the target model's performance on clean data. Smaller values indicate better adversarial transferability, and \textbf{bold} highlights the best results. Avg. is the unweighted arithmetic mean of the five reported metric scores without normalization. }%
\label{tab_trans_3}
\centering
\resizebox{0.9\linewidth}{!}{
\begin{tabular}{lcccccccccc}
\toprule
 {} & {Baseline} & {Co-Attack} & {SGA} & {VLATTACK} & {DRA} & {SA-AET} & {SimVLA} & {HQA-VLAttack} & {SADCA} & {RAO} \\ \hline
{B@4} & 39.59 & 37.27 & 34.23 & 36.23 & 33.62 & 33.37 & 34.22 & 31.80 & 31.85 & \textbf{30.83} \\
{METEOR} & 30.87 & 29.67 & 28.41 & 28.18 & 27.85 & 27.60 & 27.95 & 26.80 & 26.62 & \textbf{26.50} \\
{ROUGE-L} & 59.67 & 57.98 & 56.16 & 57.46 & 55.62 & 54.69 & 55.73 & 54.10 & 54.15 & \textbf{53.90} \\
{CIDEr} & 133.02 & 125.19 & 115.45 & 126.31 & 111.30 & 107.65 & 114.04 & 104.60 & \textbf{104.30} & 104.58 \\
{SPICE} & 23.51 & 22.58 & 21.46 & 22.49 & 20.87 & 20.52 & 21.04 & 19.80 & \textbf{19.27} & 19.47 \\
\midrule
{Avg.} & 57.33 & 54.54 & 51.14 & 54.13 & 49.85 & 48.77 & 50.60 & 47.42 & 47.24 & \textbf{47.06} \\
\bottomrule
\end{tabular}
}
\end{table*}

\begin{figure}[t]
\centering
\includegraphics[width=0.8\columnwidth]{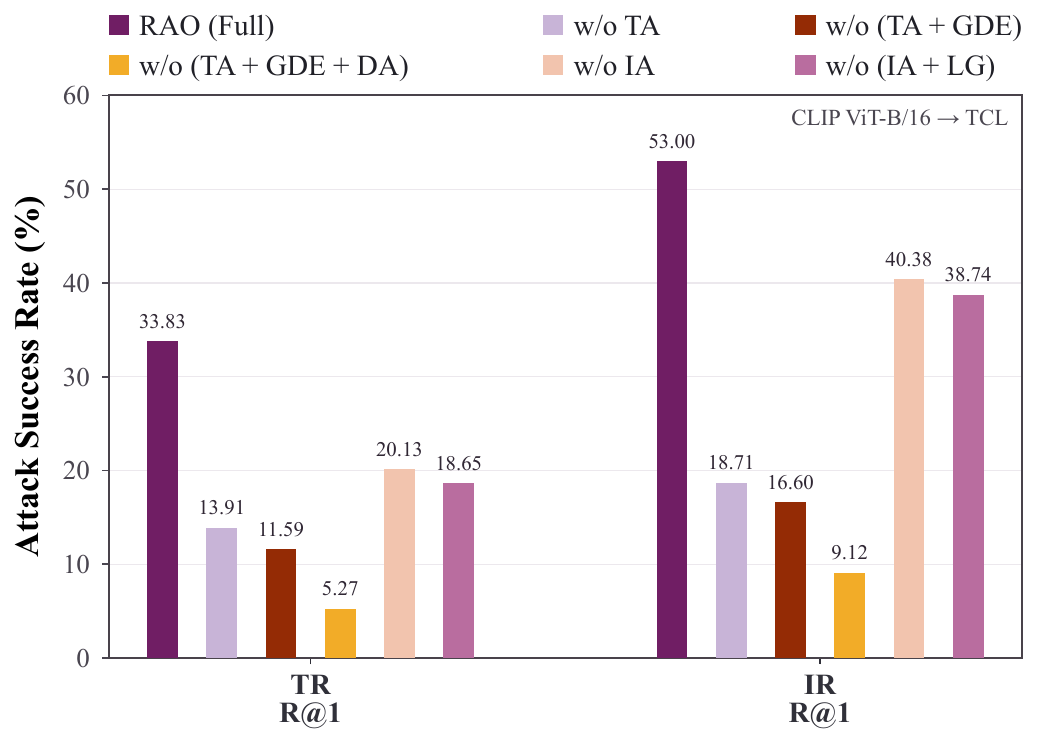}
\caption{Ablation study on Flickr30K with CLIP ViT-B/16 as the surrogate and TCL as the black-box target. We report R@1 attack success rates for text retrieval (TR) and image retrieval (IR). TA/IA denote text/image attacks; GDE, DA, and LG denote gradient direction exploration, data augmentation, and local guidance in text attacks, respectively. Higher is better.}
\label{fig:ablation}
\end{figure}

{\color{black}
To evaluate component contributions, we conduct an ablation study on Flickr30K image-text retrieval with CLIP$_{\text{ViT-B/16}}$ as the surrogate model and TCL as the black-box target. In Fig.~\ref{fig:ablation}, RAO (Full) achieves the highest R@1 ASR in TR and IR, reaching 33.83\% and 53.00\%, respectively. Removing text attack (TA) reduces these scores to 13.91\% and 18.71\%, while removing image attack (IA) yields 20.13\% and 40.38\%. These results support combining attacks in both modalities.

For image attack alone, removing gradient direction exploration (GDE) reduces TR and IR ASR from 13.91\% and 18.71\% to 11.59\% and 16.60\%, respectively. Further removing data augmentation (DA) lowers the scores to 5.27\% and 9.12\%, demonstrating the contributions of historical direction exploration and data augmentation. For text attack alone, removing local guidance (LG) reduces TR and IR ASR from 20.13\% and 40.38\% to 18.65\% and 38.74\%, respectively, supporting the use of local visual information for candidate evaluation.
\par}

\begin{figure}[!b]
\centering
{\includegraphics[angle=0, width=0.5\textwidth]{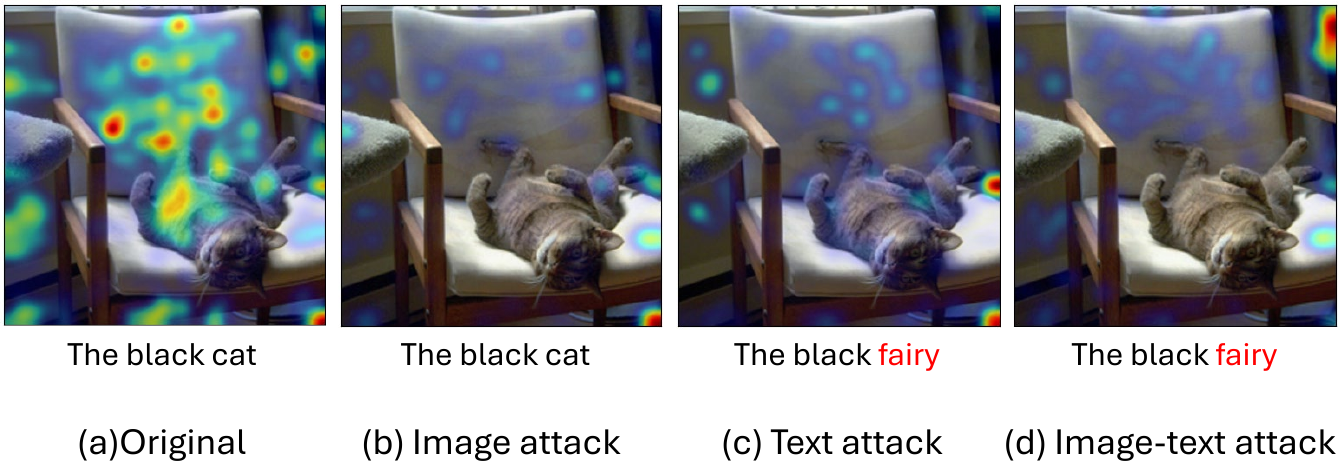}}
\caption{A Grad-CAM visualization of adversarial examples generated by the proposed method, with CLIP$_{\text{ViT-B/16}}$ and ALBEF as the source and target models, respectively. The proposed method significantly shifts the attention of the target model.}%
\label{Grad}
\end{figure}
\subsection{Visualization}
In Figure \ref{Grad}, we provide several Grad-CAM~\citep{selvaraju2017grad} visualization examples, where CLIP$_{\text{ViT-B/16}}$ and ALBEF are the source and target models. From the figure, it can be observed that the proposed method effectively shifts the model’s attention through image and/or text attacks. Furthermore, the image perturbations remain imperceptible to human observers. In contrast, adversarial attacks in text are more noticeable due to the discrete nature of language tokens. Enhancing the invisibility of text attacks remains a promising direction for future research.

\begin{figure}[!t]
\centering
\includegraphics[width=\columnwidth]{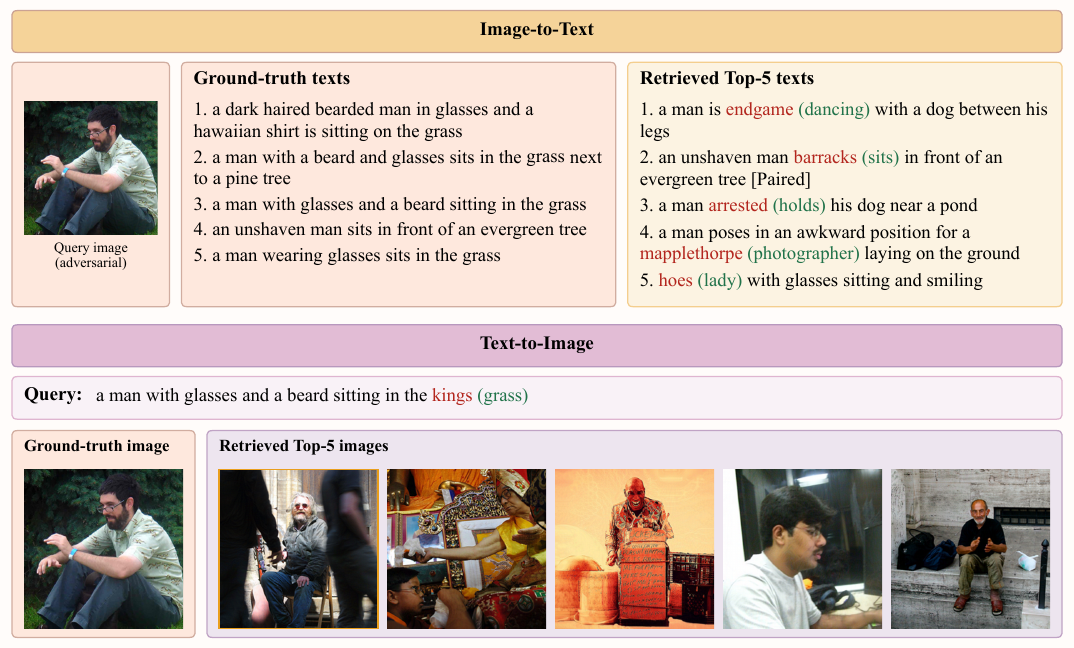}
\caption{Top-5 image-to-text (top) and text-to-image (bottom) retrieval results on Flickr30K under RAO. Red marks substituted words, and green parentheses show the original words.}
\label{fig:retrieval-case}
\end{figure}

Fig.~\ref{fig:retrieval-case} illustrates the effect of RAO on both retrieval directions. In image-to-text retrieval, the highest-ranked paired caption appears at rank 2, while in text-to-image retrieval, the paired image ranks outside the top five.

\section{Conclusions}\label{CONCLUSION}
In this paper, we identify insufficient exploration of the adversarial search space as a key factor limiting transferability in vision–language pretraining (VLP) models. To address this limitation, we propose a novel attack, termed Reorienting Adversarial Optimization (RAO), which explores a broader range of optimization directions in both image and text modalities. For images, RAO leverages historical gradient information to construct diverse candidate directions and aggregates updates from different perturbation locations to improve adversarial optimization. For text, RAO comprehensively explores candidate word substitutions across different positions, guided by both global and local cross-modal semantic discrepancies. Extensive experiments demonstrate that RAO achieves superior adversarial transferability compared with existing methods, while theoretical analysis provides further insights into perturbation interactions.

\clearpage

\bibliography{conference}
\bibliographystyle{IEEEtran}

\clearpage
\appendices
\section{Algorithm}

{\color{black}
The overall procedure is summarized in Algorithm~\ref{alg1}. First, the text attack generates candidate substitutions using Counter-Fitting with BERT as a fallback and selects adversarial texts using the global-local objective. Cross-modal Misalignment then provides the batch-level image attack objective $\mathcal J_I$. Gradient Direction Exploration combines current and historical input gradients to construct candidates in the uniform and Gaussian search spaces and retains candidates whose objective exceeds the unperturbed baseline. At the current and explored perturbation locations, sign-gradient updates are computed separately and then summed to refine the current perturbation, following Eq.~\ref{eq:image-perturbation-update}. Current-point input gradients are stored for subsequent exploration. In the algorithm, the perturbation locations $\tilde\delta\in\tilde\Delta_k$ and their updates are represented over the batch, and clipping and sign operations are applied elementwise.
\par}

\begin{algorithm*}[t!]
\color{black}
\small
\caption{Reorienting Adversarial Optimization}\label{alg1}
\KwIn{Image batch $\{x_i\}_{i=1}^{B}$ and associated texts $\{t_j\}$; matched and mismatched index sets $\mathcal P_i,\mathcal N_i$; encoders $f_I,f_T$; local region count $K$; iteration count $T$; candidate count $D$ per search space; budgets $\epsilon_I,\epsilon_T$; exploration and update step sizes $\lambda,\alpha$; CF cosine similarity threshold $\tau$.}
\KwOut{Adversarial images $\{x_i'\}$ and adversarial texts $\{t_j'\}$.}
\textbf{Stage 1: Generate adversarial texts}\;
\ForEach{image $x_i$ and its matched texts $\{t_j:j\in\mathcal P_i\}$}{
    Extract $K$ regions $\mathcal R(x_i)$ from the four corners and center\;
    \ForEach{matched text $t_j$}{
        Construct $\tilde T_j$ over valid word positions using CF neighbors with cosine similarity greater than $\tau$; use BERT only for words absent from the CF vocabulary, subject to $\epsilon_T$\;
        Select $t_j'=\arg\max_{\tilde t\in\tilde T_j}\mathcal L_{GL}(x_i,\tilde t)$ using Eq.~\ref{eq3-1}\;
    }
}
\textbf{Stage 2: Generate adversarial images}\;
Initialize $\Delta_0=0$ and an empty history of input gradients\;
Fix the original and adversarial texts, and evaluate $\mathcal J_I(\mathbf 0)$ using Eq.~\ref{eq:image-contrastive}\;
\For{$k=1,\ldots,T$}{
    Evaluate and store the current-point input gradients $g(x_i+\delta_{i,k-1})$ for all images\;
    \For{$m=1,2$}{
        Initialize $\mathbb G_k^{m+}=\varnothing$\;
        Sample $D$ coefficient sets from $\mathcal U(0,1)$ for $m=1$ or $\mathcal N(0,1)$ for $m=2$, and normalize each set so that $\sum_{q=1}^{k}\mu_q^{k,d}=1$\;
        \For{$d=1,\ldots,D$}{
            Construct $\hat g_{i,k}^{m,d}=\sum_{q=1}^{k}\mu_q^{k,d}g(x_i+\delta_{i,q-1})$ for each image using Eq.~\ref{eq:image-direction-combination}\;
            Form $\hat\Delta_k^{m,d}$ with $\hat\delta_{i,k}^{m,d}=\delta_{i,k-1}+\lambda\hat g_{i,k}^{m,d}$\;
            \If{$\mathcal J_I(\hat\Delta_k^{m,d})>\mathcal J_I(\mathbf 0)$}{
                Retain the candidate direction in $\mathbb G_k^{m+}$ according to Eq.~\ref{eq:image-direction-selection}\;
            }
        }
    }
    Form $\tilde\Delta_k$ from the current perturbation batch and the explored perturbation batches selected from the two search spaces\;
    \ForEach{$\tilde\delta\in\tilde\Delta_k$}{
        $\epsilon_k^{(\tilde\delta)}=\alpha\operatorname{sign}\big(\nabla_{\tilde\delta}\mathcal J_I(\tilde\delta)\big)$\;
    }
    $\Delta_k\leftarrow\operatorname{Clip}_{\epsilon_I}\big(\Delta_{k-1}+\sum_{\tilde\delta\in\tilde\Delta_k}\epsilon_k^{(\tilde\delta)}\big)$ using Eq.~\ref{eq:image-perturbation-update}\;
    \For{$i=1,\ldots,B$}{
        $x_{i,k}'\leftarrow\operatorname{Clip}_{[0,1]}(x_i+\delta_{i,k})$\;
    }
}
\Return{$\{x_{i,T}'\}_{i=1}^{B},\{t_j'\}$}\;
\end{algorithm*}

\section{Proof of Proposition~\ref{pro:2}} \label{sec:proof_mi}
For theoretical tractability, we consider a simplified analytical setting of RAO. Specifically, RAO combines historical gradients to construct a new exploration direction, which is then integrated with the current perturbation to generate a new perturbation. Since historical perturbations are accumulated from gradient updates, this process can be reformulated as a weighted combination of historical perturbations when the sign operation, gradient normalization, and clipping are omitted. To simplify the theoretical analysis, we further consider a uniform averaging model of historical perturbations, allowing us to derive closed-form expressions for gradient and perturbation updates. Candidate selection and gradient aggregation at multiple perturbation locations are omitted in this analysis. Specifically, the update gradient in the simplified model is:
\begin{equation} \label{eq:revised_our}
\centering
   \begin{split}
   & g_k \defeq{g(x+\frac{1}{k} \sum_{i=1}^{k} \delta_{i-1})}\\
    \end{split}
\end{equation} 
Here, $k$ denotes the iteration index. Consistent with the theoretical analysis, $g(u)=\nabla_u\ell(f_I(u),f_T(t))$ denotes the image-input gradient at $u$, with $t$ held fixed; thus, $g(x)$ is the gradient evaluated at the clean image.
The update of the perturbation in our attack at step $k$ is defined as:
\begin{equation} \label{eq:delta_our}
    \begin{split}
    &\delta_k = { \alpha \sum_{i=1}^{k}g_{i}},\\
    \end{split}
\end{equation}

\begin{proposition} \label{pro:1-1}
Applying the Taylor expansion, we can have:
\begin{equation} \label{eq:pro_1_0}
\centering
   \begin{split}
    & g(x+\delta_{i-1}) = g(x) +  H(x) \delta_{i-1} + r^1_{i-1},\\
    \end{split}
\end{equation} 
where $r^1_{i-1}$ denotes the Taylor expansion remainder.

The adversarial perturbation generated by our method via gradient ascent can then be written as: 
\begin{equation} \label{eq:pro_1_1}
    \begin{split}
    &g_k \defeq{{a_k g + \alpha b_k H g + \Tilde{R}_k^{(g)}}},\\
    &\delta_k \defeq{{c_k \alpha g + d_k \alpha^2 H g + \Tilde{R}_k^{(\delta)}}},\\
    \end{split}
\end{equation}
where 
\begin{equation} \label{eq:pro_1_2}
    \begin{split}
    &a_k = 1,\\
    &b_k = \frac{k-1}{2},\\
    &c_k = k,\\
    &d_k = \frac{(k-1)k}{4},\\
    \end{split}
\end{equation}
where $g=g(x)$ and $H=H(x)$ denote the gradient and Hessian matrix at the clean image, respectively, while $R_k^{(g)}$ and $R_k^{(\delta)}$ collect the corresponding higher-order terms.
\end{proposition}

\begin{proof}
The update gradient $g_k$ is defined as the average of historical gradients, and $\delta_k$ is obtained by accumulating these update gradients. Formally, we have:
\begin{equation}\label{eq:pro_1_3}
\begin{aligned}
g_k
&=g\!\left(x+\frac{1}{k}\sum_{i=1}^{k}\delta_{i-1}\right)\\
&=\left(g+\frac{1}{k}\sum_{i=1}^{k}
H\big(c_{i-1}\lambda g
+d_{i-1}\lambda^2Hg
+\widetilde R_{i-1}^{1}\big)
+r^1_{i-1}\right)\\
&=g+\lambda\frac{1}{k}
\left(\sum_{i=1}^{k}c_{i-1}\right)Hg
+\widetilde R_t^{2},
\end{aligned}
\end{equation}

\begin{equation}\label{eq:pro_1_4}
\begin{aligned}
\delta_k
&=\alpha\sum_{i=1}^{k}g_i\\
&=\alpha\sum_{i=1}^{k}
\left(a_i g+\alpha b_i Hg+R_i^{(g)}\right).
\end{aligned}
\end{equation}

These expressions allow us to explicitly analyze the accumulation of gradient contributions and perturbation interactions, which leads to:
\begin{equation} \label{eq:pro_1_5}
\centering
   \begin{split}
    &a_{k} = 1.\\
    \end{split}
\end{equation} 

\begin{equation} \label{eq:pro_1_6}
\centering
   \begin{split}
   &c_{k} =\sum_{j=1}^{k} a_j =k.\\
    \end{split}
\end{equation}

The coefficient of $Hg$, i.e., $b_k$, can be obtained as follows:
\begin{equation} \label{eq:pro_1_7}
\centering
   \begin{split}
   &b_k = \sum_{i=1}^{k} \mu_{i}^k c_{i-1}.\\
    \end{split}
\end{equation} 

\begin{equation} \label{eq:pro_1_8}
\centering
   \begin{split}
   &d_k = \sum_{i=1}^{k}  b_{i}\\
   &=\sum_{i=1}^{k} \sum_{j=1}^{i} \mu_{j}^i c_{j-1}.\\
    \end{split}
\end{equation}

At the same time, we have:
\begin{equation}\label{eq:delta-recurrence}
\begin{aligned}
\delta_{k+1}
&=\delta_k+\alpha g_{k+1}\\
&=\left(c_k\alpha g+d_k\alpha^2Hg
+R_k^{(\delta)}\right)\\
&\quad+\alpha\left(
a_{k+1}g+\alpha b_{k+1}Hg
+R_{k+1}^{(g)}\right)\\
&=(c_k+a_{k+1})\alpha g\\
&\quad+(d_k+b_{k+1})\alpha^2Hg
+R_{k+1}^{(\delta)}.
\end{aligned}
\end{equation}

Thus, the coefficients update recursively as:
\begin{equation}
\begin{aligned}
c_k+a_{k+1}
&=k+1\\
&=c_{k+1},
\end{aligned}
\end{equation}

\begin{equation}
\centering
   \begin{split}
   &d_k + b_{k+1} =  \frac{k(k-1)}{4} + \frac{k}{2}\\
   & = \frac{k(k+1)}{4}\\
   & = d_{k+1}.\\
    \end{split}
\end{equation} 

According to Section \ref{Theo_anan}, the Shapley interaction between perturbation units $a, b$ in $\delta_k$ is calculated as follows:
\begin{equation}\label{eq:mi_interaction}
\begin{aligned}
I_{ab}(\delta_k)
&=\delta_{k,a}H_{ab}\delta_{k,b}
+\widetilde R_2(\delta_k)\\
&=\left(c_k\alpha g_a
+d_k\alpha^2 g^TH_{*a}\right)H_{ab}\\
&\quad\left(c_k\alpha g_b
+d_k\alpha^2 g^TH_{*b}\right)
+\mathcal R_k\\
&=c_k^2\alpha^2g_ag_bH_{ab}\\
&\quad+c_kd_k\alpha^3g_bH_{ab}g^TH_{*a}\\
&\quad+c_kd_k\alpha^3g_aH_{ab}g^TH_{*b}
+\mathcal R_k'\\
&=k^2\alpha^2g_ag_bH_{ab}\\
&\quad+\frac{k^2(k-1)}{4}
\alpha^3g_bH_{ab}g^TH_{*a}\\
&\quad+\frac{k^2(k-1)}{4}
\alpha^3g_aH_{ab}g^TH_{*b}
+\mathcal R_k',
\end{aligned}
\end{equation}

Here, $O(H^2)$ denotes second-order small terms with respect to $H$, which are neglected in the following calculations. According to~\citep{wang2020unified}, we have $\mathbb{E}_{a, b} [g_a g_b g^T H{*b}] \geq 0$. The interaction for perturbations learned by a general multi-step attack, e.g., SGA, is given by $I_{ab}(\delta^{\textrm{msa}}_k) = k^2 \alpha^2 g_a g_b H_{ab} + \frac{k^2(k-1)}{2} \alpha^3 g_b H_{ab} g^T H_{*a} + \frac{k^2(k-1)}{2} \alpha^3 g_a H_{ab} g^T H_{*b} + O(H^2).$ As a result, $\mathbb{E}_{a, b}(I_{a b}(\delta_k) - I_{a b}(\delta^{\textrm{msa}}_k)) = \mathbb{E}_{a, b}(- \frac{k^2(k-1)}{4} \alpha^3 g_b H_{ab} g^T H_{*a} -  \frac{k^2(k-1)}{4}  \alpha^3 g_a H_{ab} g^T H_{*b}) <0$.
Therefore, compared to the general multi-step attack, the proposed method induces fewer interactions, thereby promoting transferability. 
\end{proof}

\end{document}